\documentclass{article} 
\usepackage{iclr2017_conference,times}
\usepackage{hyperref}
\usepackage{url}

\usepackage{graphicx}
\usepackage{algorithm}
\usepackage{algorithmic}
\usepackage{amsmath}
\usepackage{amssymb}
\usepackage{graphicx}
\usepackage{caption}
\usepackage{subcaption}
\usepackage{multirow}
\usepackage{wasysym}
\usepackage{makecell}

\graphicspath{{figures/}}

\title{Neural Map: Structured Memory for Deep Reinforcement Learning}

\author{Emilio Parisotto \& Ruslan Salakhutdinov \\
Department of Machine Learning\\
Carnegie Mellon University \\
Pittsburgh, PA 15213, USA \\
\texttt{\{eparisot,rsalakhu\}@cs.cmu.edu} \\
}

\begin{document}

\maketitle

\begin{abstract}

A critical component to enabling intelligent reasoning in partially observable environments is memory.
Despite this importance, Deep Reinforcement Learning (DRL) agents have so far used relatively simple memory architectures, with the main methods to overcome partial observability being either a temporal convolution over the past $k$ frames or an LSTM layer.
More recent work~\citep{MinecraftMemnet16} has went beyond these architectures by using memory networks which can allow more sophisticated addressing schemes over the past $k$ frames. But even these architectures are unsatisfactory due to the reason that they are limited to only remembering information from the last $k$ frames. 
In this paper, we develop a memory system with an adaptable write operator that is customized to the sorts of 3D environments that DRL agents typically interact with. This architecture, called the Neural Map, uses a spatially structured 2D memory image to learn to store arbitrary information about the environment over long time lags. We demonstrate empirically that the Neural Map surpasses previous DRL memories on a set of challenging 2D and 3D maze environments and show that it is capable of generalizing to environments that were not seen during training.

\vspace{-0.1in}

\end{abstract}

\section{Introduction}

Memory is a crucial aspect of an intelligent agent's ability to plan and reason in partially observable environments.
Without memory, agents must act reflexively according only to their immediate percepts and cannot execute plans that occur over an extended time interval.
Recently, Deep Reinforcement Learning agents have been capable of solving many challenging tasks such as Atari Arcade Games~\citep{Atari15}, robot control~\citep{levine2016end}  and 3D games such as Doom~\citep{DeepDoom16}, but successful behaviours in these tasks have often only been based on a relatively short-term temporal context or even just a single frame. On the other hand, many tasks require long-term planning, such as a robot gathering objects or an agent searching a level to find a key in a role-playing game.

Neural networks that utilized external memories have recently had an explosion in variety, which can be distinguished along two main axis: memories with write operators and those without.
Writeless external memory systems, often referred to as ``Memory Networks'' \citep{EndToEndMemnet15, MinecraftMemnet16}, typically fix which memories are stored. For example, at each time step, the memory network would store the past M states seen in an environment. What is learnt by the network is therefore how to access or read from this fixed memory pool, rather than what contents to store within it.

The memory network approach has been successful in language modeling, question answering \citep{EndToEndMemnet15} and was shown to be a sucessful memory for deep reinforcement learning agents in complex 3D environments \citep{MinecraftMemnet16}.
By side-steping the difficulty involved in learning what information is salient enough to store in memory, the memory network introduces two main disadvantages. The first disadvantage is that a potentially significant amount of redundant information could be stored. The second disadvantage is that a domain expert must choose what to store in the memory, e.g. for the DRL agent, the expert must set M to a value that is larger than the time horizon of the currently considered task.

On the other hand, external neural memories having write operations are potentially far more efficient, since they can learn to store salient information for unbounded time steps and ignore any other useless information, without explicitly needing any a priori knowledge on what to store. 
One prominent research direction within write-based architectures has been neural memories based on the types of memory structures that are found in computers, such as tapes~\citep{NTM14}, RAM~\citep{kurach2016nram}, and GPUs~\citep{kaiser2016ngpu}. In contrast to typical recurrent neural networks, these neural computer emulators have far more structured memories which follow many of the same design paradigms that digital computers have traditionally utilized. One such model, the Differentiable Neural Computer (DNC) \citep{DNC16} and its predecessor the Neural Turing Machine (NTM) \citep{NTM14}, structure the architecture to explicitly separate memory from computation.
The DNC has a recurrent neural controller that can access an external memory resource by executing differentiable read and write operations. This allows the DNC to act and memorize in a structured manner resembling a computer processor, where read and write operations are sequential and data is store distinctly from computation. The DNC has been used sucessfully to solve complicated algorithmic tasks, such as finding shortest paths in a graph or querying a database for entity relations.

Building off these previous external memories, we introduce a new architecture called the Neural Map, a structured memory designed specifically for reinforcement learning agents in 3D environments. The Neural Map architecture overcomes some of the shortcomings of the previously mentioned neural memories. First, it uses an adaptable write operation and so its size and computational cost does not grow with the time horizon of the environment as it does with memory networks. Second, we impose a particular inductive bias on the write operation so that it is 1) well suited to 3D environments where navigation is a core component of sucessful behaviours, and 2) uses a sparse write operation that prevents frequent overwriting of memory locations that can occur with NTMs and DNCs. To accomplish this, we structure a DNC-style external memory in the form of a 2-dimensional map, where each position in the map is a distinct memory.

To demonstrate the effectiveness of the neural map, we run it in on variety of 2D partially-observable maze-based environments and test it against LSTM and memory network policies. Finally, to establish its scalability, we run a Neural Map agent on a challenging 3D maze environment based on the video game Doom.

\section{Background}

A Markov Decision Process (MDP) is defined as a tuple $(\mathcal{S}, \mathcal{A}, \mathcal{T}, \gamma, \mathcal{R})$ where
$\mathcal{S}$ is a finite set of states,
$\mathcal{A}$ is a finite set of actions,
$\mathcal{T}(s'|s,a)$ is the transition probability of arriving in state $s'$ when executing action $a$ in initial state $s$,
$\gamma$ is a discount factor, 
and $\mathcal{R}(s,a,s')$ is the reward function of executing action $a$ in state $s$ and ending up at state $s'$.
We define a policy $\pi(\cdot|s)$ as a mapping from a state $s$ to a distribution over actions, where $\pi(a_i|s)$ denotes the probability of action $a_i$ given that we are in state $s$. The value of a policy $V^\pi(s)$ is the expected discounted cumulative reward when starting from state $s$ and sampling actions according to $\pi$, i.e.:
\begin{align}
  V^\pi(s) = \mathbb{E}_\pi\left[ \sum_{t=0}^\infty \gamma^t R_t | s_0 = s\right] 
\end{align}

An optimal value function, denoted $V^*(s)$, is the maximum value we can get from state $s$ according to any policy, i.e. $V^*(s) = \max_\pi V^\pi(s)$. An optimal policy $\pi^*$ is defined as a policy which achieves optimal value at each state, i.e. $V^{\pi^*}(s) = V^*(s)$. An optimal policy is guaranteed to exist~\citep{SuttonBarto98}.

The REINFORCE algorithm~\citep{REINFORCE92} iteratively updates a given policy $\pi$ in the direction of the optimal policy. This update direction is defined by $\nabla_\pi \log \pi(a_t|s_t)G_t$
with $G_t = \sum_{k=0}^\infty \gamma^k R_{t+k}$ being the future cumulated reward for a particular episode rollout. The variance of this update is typically high but can be reduced by using a ``baseline'' $b_t(s_t)$, which is a function of the current state. Therefore the baseline-augmented update equation is $\nabla_\pi \log \pi(a_t|s_t)(G_t - b_t(s_t))$. The typically used baseline is the value function, $b_t(s_t) = V^\pi(s_t)$. This combination of REINFORCE with value function baseline is commonly termed the ``Actor-Critic'' algorithm.

In this paper, we utilize a modified Asynchronous Advantage Actor-Critic (A3C)~\citep{A3C16}, which can be seen as a specialization of the actor-critic framework when using deep networks to parameterize the policy and value function. The policy is a function of the state, parameterized as a deep neural network:
  $\pi(a|s) = f_\theta(s, a)$,
where f is a deep neural network with parameter vector $\theta$.

We make a modification to the A3C framework where instead of executing and updating a policy asynchronously in several concurrent environments and then occasionally synchronizing network parameters after several updates, we execute and update the policy synchronously at each time step across several concurrent environments, meaning that the network parameters never require synchronization. Other than this modification, the algorithm is identical to the one used in A3C, including the methods used to update the value and policy networks (refer to ~\citep{A3C16} for details). 

\section{Neural Map}

In this section, we will describe the details of the neural map. We assume we want our agent to act within some 2- or 3-dimensional environment. The neural map is the agent's internal memory storage that can be read from and written to during interaction with its environment, but where the write operator is selectively limited to affect only the part of the neural map that represents the area where the agent is currently located. For this paper, we assume for simplicity that we are dealing with a 2-dimensional map. This can easily be extended to 3-dimensional or even higher-dimensional maps (i.e. a 4D map with a 3D sub-map for each cardinal direction the agent can face).

Let the agent's position be $(x,y)$ with $x \in\mathbb{R}$ and $y \in\mathbb{R}$ and let the neural map $M$ be a $C\times H\times W$ feature block, where $C$ is the feature dimension, $H$ is the vertical extent of the map and $W$ is the horizontal extent. Assume there exists some coordinate normalization function $\psi(x,y)$ such that every unique $(x,y)$ can be mapped into $(x',y')$, where $x'\in \{0,\hdots,W\}$ and $y'\in \{0,\hdots,H\}$. For ease of notation, suppose in the sequel that all coordinates have been normalized by $\psi$ into neural map space.

Let $s_t$ be the current state embedding, $M_t$ be the current neural map, and $(x_t,y_t)$ be the current position of the agent within the neural map. 
The Neural Map is defined by the following set of equations:
\begin{align}
  r_t                 &= read(M_t) \\
  c_t                 &= context(M_t, s_t, r_t) \\
  w_{t+1}^{(x_t,y_t)} &= write(s_t, r_t, c_t, M_t^{(x_t,y_t)}) \\
  M_{t+1}             &= update(M_t, w_{t+1}^{(x_t,y_t)}) \\
  o_t                 &= [r_t, c_t, w_{t+1}^{(x_t,y_t)}] \\
  \pi_t(a|s)          &= \text{Softmax}(f(o_t)),
\end{align}
where 
$w_{t}^{(x_t,y_t)}$ represents the feature at position $(x_t,y_t)$ at time $t$,
$[x_1,\hdots,x_k]$ represents a concatenation operation, and 
$o_{t}$ is the output of the neural map at time $t$ which is then processed by another deep network $f$ to get the policy outputs $\pi_t(a|s)$. We will now separately describe each of the above operations in more detail.

\begin{figure}[t]
  \centering
  \includegraphics[width=0.5\linewidth]{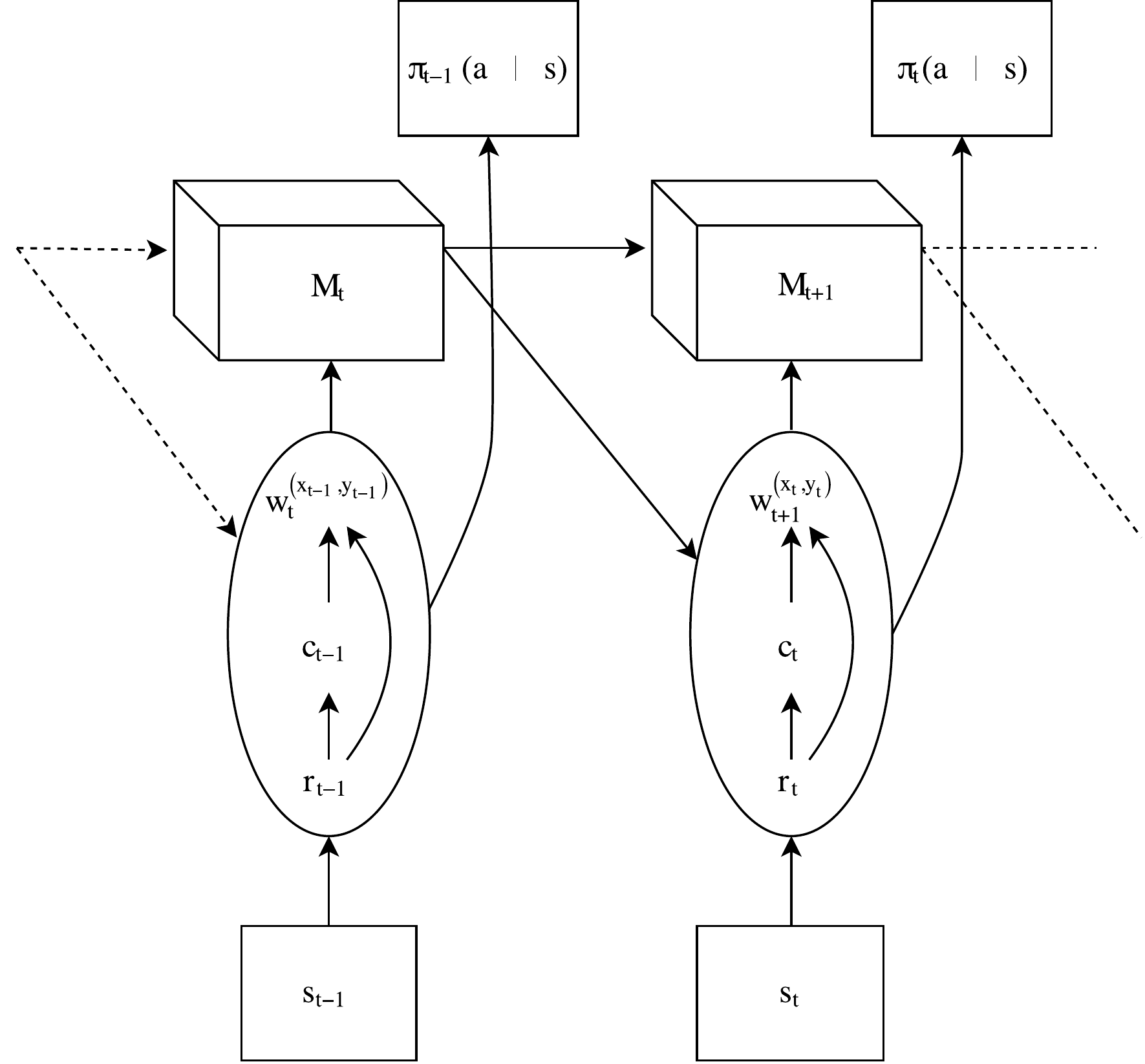}
  \caption{\small A visualization of two time steps of the neural map.}
\vspace{-0.1in}
\end{figure}

\subsection{Global Read Operation}

The $read$ operation passes the current neural map $M_t$ through a deep convolutional network and produces a $C$-dimensional feature vector $r_t$. The global read vector $r_t$ summarizes information about the entire map.

\subsection{Context Read Operation}

The $context$ operation performs context-based addressing to check whether certain features are stored in the map. It takes as input the current state embedding $s_t$ and the current global read vector $r_t$ and first produces a query vector $q_t$. The inner product of the query vector and each feature $M_t^{(x,y)}$ in the neural map is then taken to get scores $a_t^{(x,y)}$ at all positions $(x,y)$. The scores are then normalized to get a probability distribution $\alpha_t^{(x,y)}$ over every position in the map, also known as 
``soft attention''~\citep{bahdanau2015nmt}. This probability distribution is used to compute a weighted average $c_t$ over all features $M_t^{(x,y)}$. To summarize: 
\begin{align}
  \label{eq:context1}
  q_t              &= W [s_t, r_t] \\
  \label{eq:context2}
  a_t^{(x,y)}      &= q_t \cdot M_t^{(x,y)} \\
  \label{eq:contextprob}
  \alpha_t^{(x,y)} &= \frac{e^{a_t^{(x,y)}}}{\sum_{(w,z)} e^{a_t^{(w,z)}}} \\
  \label{eq:context3}
  c_t              &= \sum_{(x,y)} \alpha_t^{(x,y)} M_t^{(x,y)}, 
\end{align}
where $W$ is a weight matrix. The context read operation allows the neural map to operate as an associative memory: the agent provides some possibly incomplete memory (the query vector $q_t$) and the operation will return the completed memory that most closely matches $q_t$. So, for example, the agent can query whether it has seen something similar to a particular landmark that is currently within its view.

\subsection{Local Write Operation}

Given the agent's current position $(x_t,y_t)$ at time $t$, the $write$ operation takes as input the current state embedding $s_t$, the global read output $r_t$, the context read vector $c_t$ and the current feature at position $(x_t,y_t)$ in the neural map $M_t^{(x_t,y_t)}$ and produces, using a deep neural network $f$, a new C-dimensional vector $w_{t+1}^{(x_t,y_t)}$. This vector functions as the new local write candidate vector at the current position $(x_t,y_t)$: 
\begin{align}
  w_{t+1}^{(x_t,y_t)} &= f([s_t, r_t, c_t, M_t^{(x_t,y_t)}])
\end{align}
\subsection{Map Update Operation}

The $update$ operation creates the neural map for the next time step. The new neural map $M_{t+1}$ is equal to the old neural map $M_t$, except at the current agent position $(x_t,y_t)$, where the current write candidate vector $w_{t+1}^{(x_t,y_t)}$ is stored:
\begin{align}
  M_{t+1}^{(a,b)} = \left\{ \begin{array}{lr}
    w_{t+1}^{(x_t,y_t)}, & \text{for } (a,b)=(x_t,y_t) \\
    M_t^{(a,b)},         & \text{for } (a,b)\neq(x_t,y_t)
  \end{array}\right.
\end{align}
\subsection{Operation Variants}

There are several modifications that can be made to the standard operations as defined above. Below we discuss some variants.

\subsubsection{Localized Read Operation}

Instead of passing the entire neural map through a deep convolutional network, a spatial subset of the map can be passed instead. For example, a Spatial Transformer Network~\citep{STN15} can be used to attentively subsample the neural map at particular locations and scales. This can be helpful when the environment requires a large high-resolution map which can be computationally expensive to process in its entirety at each time step.

\subsubsection{Key-Value Context Read Operation}

We can impose a stronger bias on the context addressing operation by splitting each feature of the neural map into two parts $M_t^{(x,y)} = [k_t^{(x,y)}, v_t^{(x,y)}]$, where $k_t^{(x,y)}$ is the $(C/2)$-dimensional ``key'' feature and $v_t^{(x,y)}$ is the $(C/2)$-dimensional ``value'' feature~\cite{key_value}. The key features are matched against the query vector (which is now a $(C/2)$-dimensional vector) to get the probability distribution $\alpha_t^{(x,y)}$, and the weighted average is taken over the value features. Concretely:
\begin{align}
  q_t              &= W [s_t, r_t] \\
  M_t^{(x,y)}      &= [k_t^{(x,y)}, v_t^{(x,y)}] \\
  a_t^{(x,y)}      &= q_t \cdot k_t^{(x,y)} \\
  \alpha_t^{(x,y)} &= \frac{e^{a_t^{(x,y)}}}{\sum_{(w,z)} e^{a_t^{(w,z)}}} \\
  c_t              &= \sum_{(x,y)} \alpha_t^{(x,y)} v_t^{(x,y)}
\end{align}
Having distinct key-value features allows the network to more explicitly separate the addressing feature space from the content feature space. 

\subsubsection{GRU-based Local Write Operation}

As previously defined, the write operation simply replaces the vector at the agent's current position with a new feature produced by a deep network. Instead of this hard rewrite of the current position's feature vector, we can use a gated write operation based on the recurrent update equations of the Gated Recurrent Unit (GRU) \citep{GRU14}. Gated write operations have a long history in unstructured recurrent networks and they have shown a superior ability to maintain information over long time lags versus ungated networks. The GRU-based write operation is defined as:
\begin{align*}
  r_{t+1}^{(x_t,y_t)}       &= \sigma(W_r [s_t, r_t, c_t, M_t^{(x_t,y_t)}]) \\
  \hat{w}_{t+1}^{(x_t,y_t)} &= \tanh(W_{\hat{h}} [s_t, r_t, c_t] + U_{\hat{h}}(r_{t+1}^{(x_t,y_t)} \odot M_t^{(x_t,y_t)})) \\
  z_{t+1}^{(x_t,y_t)}       &= \sigma(W_z [s_t, r_t, c_t, M_t^{(x_t,y_t)}]) \\
  w_{t+1}^{(x_t,y_t)}       &= (1 - z_{t+1}^{(x_t,y_t)}) \odot M_t^{(x_t,y_t)} + z_{t+1}^{(x_t,y_t)} \odot \hat{w}_{t+1}^{(x_t,y_t)},
\end{align*}
where $x \odot y$ is the Hadamard product between vectors $x$ and $y$, $\sigma(\cdot)$ is the sigmoid activation function and $W_*$ and $U_*$ are weight matrices. Using GRU terminology, $r_{t+1}^{(x_t,y_t)}$ is the reset gate, $\hat{w}_{t+1}^{(x_t,y_t)}$ is the candidate activation and $z_{t+1}^{(x_t,y_t)}$ is the update gate.
By making use of the reset and update gates, the GRU-based update can modulate how much the new write vector should differ from the currently stored feature.

\section{Experiments}

To demonstrate the effectiveness of the Neural Map, we run it on 2D and 3D maze-based environments where memory is crucial to optimal behaviour. We compare to previous memory-based DRL agents, namely a simple LSTM-based agent which consists of a single pre-output LSTM layer as well as MemNN~\citep{MinecraftMemnet16} agents. Of the agents presented in \cite{MinecraftMemnet16}, we use the MQN version, i.e. the stand-alone memory network without an LSTM layer.

\begin{table*}
\small
  \centering
  \begin{tabular}{|c|ccc|ccc|}
    \hline
    \multirow{4}{*}{Agent}
    & \multicolumn{6}{|c|}{Goal-Search} \\ \cline{2-7}
    & \multicolumn{3}{|c|}{Train} & \multicolumn{3}{|c|}{Test} \\
    & 7-11 & 13-15 & Total & 7-11 & 13-15 & Total \\ \hline
    Random     & 41.9\% & 25.7\% & 38.1\% & 46.0\% & 29.6\% & 38.8\% \\ \hline
    LSTM       & 60.6\% & 41.8\% & 59.3\% & 65.5\% & 47.5\% & 57.4\% \\ \hline
    MemNN-32   & 85.1\% & 58.2\% & 77.8\% & 92.6\% & 69.7\% & 83.4\% \\ \hline
    Neural Map & 92.4\% & 80.5\% & 89.2\% & 93.5\% & 87.9\% & 91.7\% \\ \hline
    \makecell{Neural Map (GRU)} & {\bf 97.0\%} & {\bf 89.2\%} & {\bf 94.9\%} & {\bf 97.7\%} & {\bf 94.0\%} & {\bf 96.4\%} \\ \hline
  \end{tabular}
  \caption{\small Results of several different agent architectures on the ``Goal-Search'' environment. The ``train'' columns represents the number of mazes solved when sampling from the same distribution as used during training. The ``test'' columns represents the number of mazes solved when run on a set of held-out maze samples which are guaranteed not to have been sampled during training. }
  \label{tab:randmazeresults}
  \vspace{-0.1in}
\end{table*}

\subsection{2D Goal-Search Environment}

\begin{figure}[t]
  \centering
  \begin{subfigure}[c]{0.3\linewidth}
    \centering
    \includegraphics[width=\linewidth]{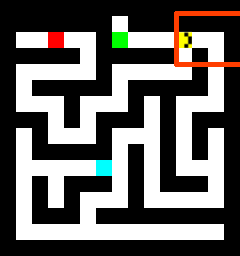}
    \caption{Maze}
    \label{fig:randmaze}
  \end{subfigure}%
  \begin{subfigure}[c]{0.2\linewidth}
    \centering
    \includegraphics[width=0.4\linewidth]{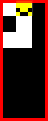}
    \caption{Observation}
    \label{fig:randmaze_obs}
  \end{subfigure}
  \caption{\small Images showing the 2D maze environments. The left side (Fig.~\ref{fig:randmaze}) represents the fully observable maze while the right side (Fig.~\ref{fig:randmaze_obs}) represents the agent observations. The agent is represented by the yellow pixel with its orientation indicated by the black arrow within the yellow block. The starting position is always the topmost position of the maze. The red bounding box represents the area of the maze that is subsampled for the agent observation. In ``Goal-Search'', the goal of the agent is to find a certain color block (either red or teal), where the correct color is provided by an indicator (either green or blue). This indicator has a fixed position near the start position of the agent.
  }
  \vspace{-0.1in}
\end{figure}

The ``Goal-Search'' environment is adapted from \cite{MinecraftMemnet16}. Here the agent starts in a fixed starting position within some randomly generated maze with two randomly positioned goal states. It then observes an indicator at a fixed position near the starting state (i.e. the green tile at the top of the maze in Fig.~\ref{fig:randmaze}). This indicator will tell the agent which of the two goals it needs to go to (blue indicator$\rightarrow$teal goal, green indicator$\rightarrow$red goal). If the agent goes to the correct goal, it gains a positive reward while if it goes to the incorrect goal it gains a negative reward. Therefore the agent needs to remember the indicator as it searches for the correct goal state.

The mazes during training are generated using a random generator. A held-out set of 1000 random mazes is kept for testing. This test set therefore represents maze geometries that have never been seen during training, and measure the agent's ability to generalize to new environments. The mazes sampled during training range from a size of $5\times 5$ to $15\times15$. In the maze generation process, we first sample maze sizes uniformly and then generate the maze. Sampling different maze sizes from easy to difficult during training is similar to a style of curriculum learning. The episode is terminated early if the agent goes 100 steps without reaching a goal.

The agent's state observations are a $5\times15\times3$ subsample of the complete maze so that the agent is able to see 15 pixel forward and 3 pixels on the side (center pixel + one pixel on each side of the agent) which is depicted in Fig.\ref{fig:randmaze_obs}. This view is obscured so the agent is prevented from seeing the identity of anything behind walls. The 5 binary channels in the observation represent object identities: channel 1 represents presence of walls, 2 represents the green indicator, 3 the blue indicator, 4 the red goal, and 5 the teal goal. 

The first baseline agent we evaluate is a recurrent network with 128 LSTM units. The other baseline is the MQN, which is a memory-network-based architecture that performs attention over the past 32 states it has seen. In more detail, the MQN stores all previous 32 states in memory and passes them all through a deep network to get an embedding, then performs a context-based lookup on this memory pool using a query vector computed from the current state. Finally, we test two Neural Map architectures, one with the standard update and one with the GRU-based update. The Neural Map architectures have an internal map size of $15\times15$ with a feature channel size of $32$. To get $r_t$, the global read operation passes the neural map first through a 3-layer convolutional network, with each convolution having filter size $3\times3$ and $8$ channels, followed by a $256$ unit linear layer and then a final $32$ unit linear layer. Both Neural Maps are identical minus the difference in write operations. All agents are trained using a synchronized Advantage Actor-Critic with 16 concurrent environments trained for 10 million steps per environment (160 million total).

The results are reported in Table~\ref{tab:randmazeresults}. During testing, we extend the maximum episode length from 100 to 500 steps so that the agent is given more time to solve the maze. From the results we can see that the Neural Map solves the most mazes in both the training and test distributions.
The results show that the Neural Map based architectures can better succeed at finding the correct goal over all other methods. In particular, the GRU-based Neural Map solves almost all of the train/test mazes. One thing to note is that the accuracy on the training distribution is slightly lower than the test set. This is because the training set encompases almost all random mazes except the 1000 of the test set thus it is likely that the agent sees each training map only once.

Beyond train/test splits, the results are further separated by maze size, which will give an idea of whether the agent is limited by the amount of time it can store information since larger mazes will require remembering information over longer time steps.
We split the 1000 test mazes into 572 small mazes (sizes between $7\times7$ to $11\times11$) and 428 large mazes (sizes between $13\times13$ to $15\times15$). Table~\ref{tab:randmazeresults} shows that the memory network and LSTM agents have significant difficulty learning how to solve longer maze sizes. On the other hand, the neural map with either standard or GRU-based updates is capable of solving larger maze sizes at a much higher rate.

\begin{figure}[t]
\vspace{-0.1in}
  \centering
  \includegraphics[width=0.5\linewidth]{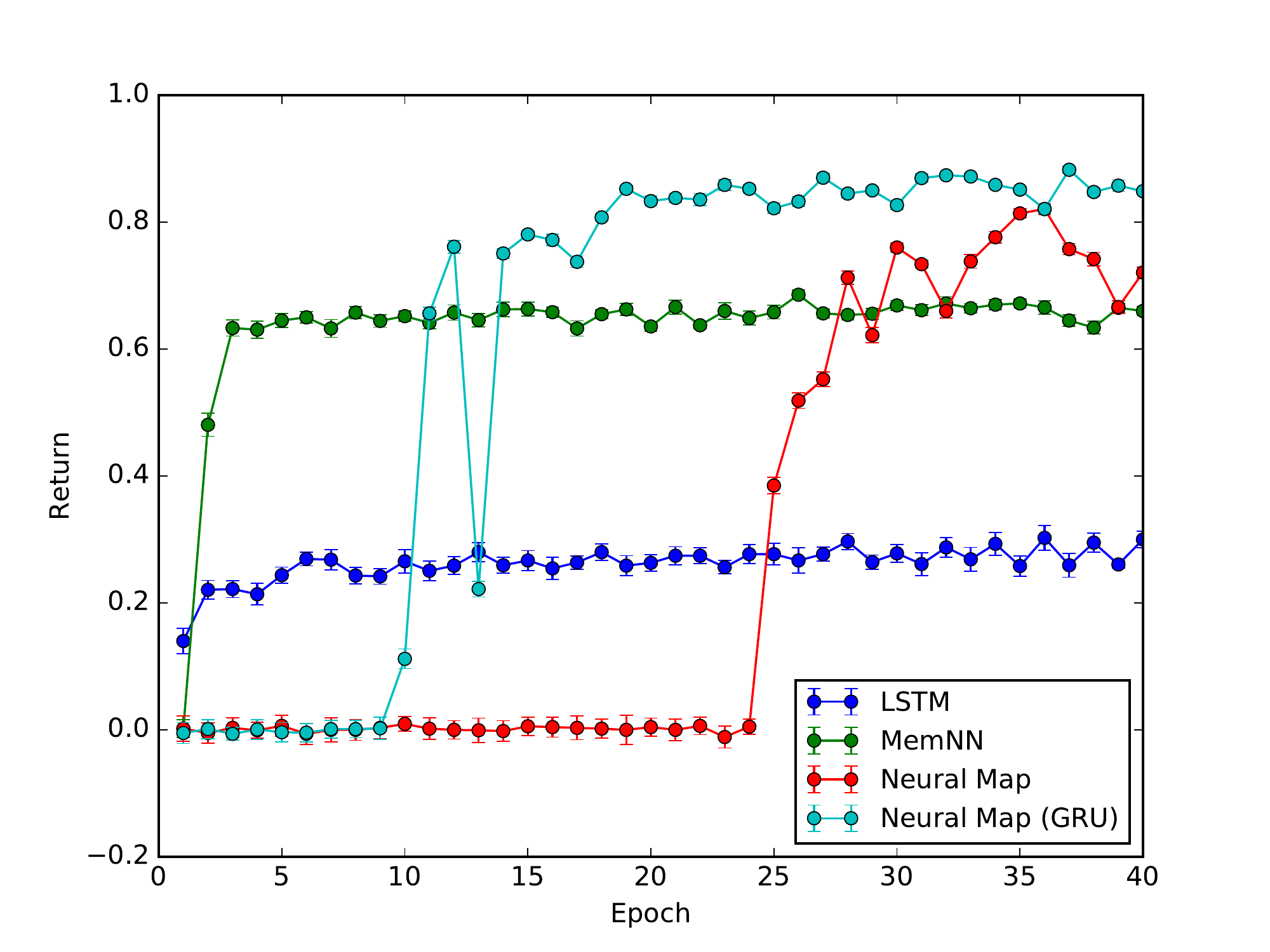}
  \vspace{-0.2in}
  \caption{\small Training curves for all 4 agent architectures on the ``Goal-Search'' environment. The x-axis is an epoch (250k concurrent steps) and the y-axis is the average undiscounted episode return. The curves show that the GRU-based Neural Map learns faster and is more stable than the standard update Neural Map.}
  \label{fig:randmazecurve}
\vspace{-0.1in}
\end{figure}

\begin{figure}[t]
  \centering
  \begin{minipage}[c]{0.125\linewidth}
    \centering
    \includegraphics[width=0.375\linewidth]{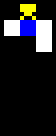} \\
    \includegraphics[width=\linewidth]{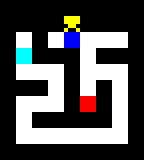} \\
    \includegraphics[width=\linewidth]{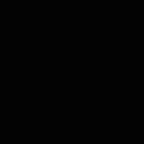}
  \end{minipage}%
  \begin{minipage}[c]{0.125\linewidth}
    \centering
    \includegraphics[width=0.375\linewidth]{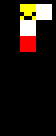} \\
    \includegraphics[width=\linewidth]{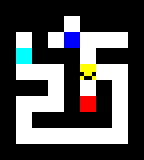} \\
    \includegraphics[width=\linewidth]{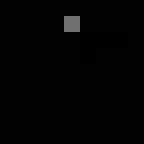}
  \end{minipage}%
  \begin{minipage}[c]{0.125\linewidth}
    \centering
    \includegraphics[width=0.375\linewidth]{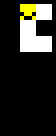} \\
    \includegraphics[width=\linewidth]{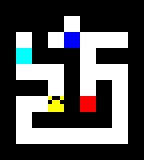} \\
    \includegraphics[width=\linewidth]{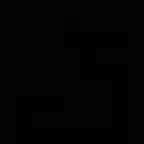}
  \end{minipage}%
  \begin{minipage}[c]{0.125\linewidth}
    \centering
    \includegraphics[width=0.375\linewidth]{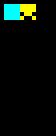} \\
    \includegraphics[width=\linewidth]{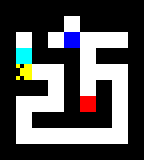} \\
    \includegraphics[width=\linewidth]{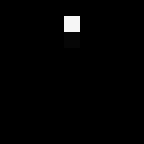}
  \end{minipage}
  \caption{\small A few sampled states from an example episode demonstrating how the agent learns to use the context addressing operation of the Neural Map. The top row of images is the observations made by the agent, the center is the fully observable mazes and the bottom image is the probability distributions over locations induced by the context operation at that step.} 
  \label{fig:contextdemo}
  \vspace{-0.1in}
\end{figure}

The training curves are plotted in Fig.~\ref{fig:randmazecurve}. We can see that the Neural Map agents get the highest final reward, but both initially learn slower than the LSTM and the MemNN. It is not surprising that the MemNN learns faster because it is a feedforward network that does not need to maintain a recurrent state. The LSTM initially learns quickly but plateaus at a pretty low average reward. Results in Table~\ref{tab:randmazeresults} suggest  that this relatively higher initial score might be due to the LSTM quickly learning how to solve the small mazes. For the Neural Maps, the GRU-based update was observed to learn much faster as well as surpass the final score of the standard update. Another benefit of the GRU-based write operation was that it typically made the Neural Map much more stable during training.

To gain some insight into what the neural map learned to do internally, we ran it on an example maze shown in Figure~\ref{fig:contextdemo}. In this figure, the top row of images are the agent observations, the center row are the fully observable mazes and the bottom row are the probability distributions over locations from the context operation, e.g. the $\alpha_t^{(x,y)}$ values defined by Eq.~\ref{eq:contextprob}. In this maze, the indicator is blue, which indicates that the teal goal should be visited. We can see that once the agent sees the incorrect red goal, the context distribution faintly focuses on the map location where the agent had observed the indicator. On the other hand, when the agent first observes the correct teal goal, the location where the agent observed the indicator lights up brightly. This means that the agent is using its context retrieval operation to keep track of the landmark (the indicator) that it has previously seen.

\subsection{3D Doom Environment}

\begin{figure}[t]
  \captionsetup[subfigure]{justification=centering}
  \centering
  \begin{subfigure}{0.7\linewidth}
    \centering
    $\vcenter{\hbox{\includegraphics[width=0.33\linewidth]{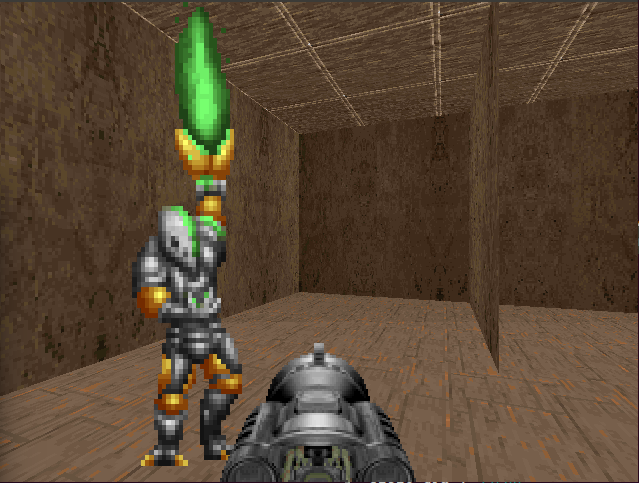}}}$ %
    $\vcenter{\hbox{\scalebox{2}{\Huge\pointer}}}$ %
    $\vcenter{\hbox{\includegraphics[width=0.33\linewidth]{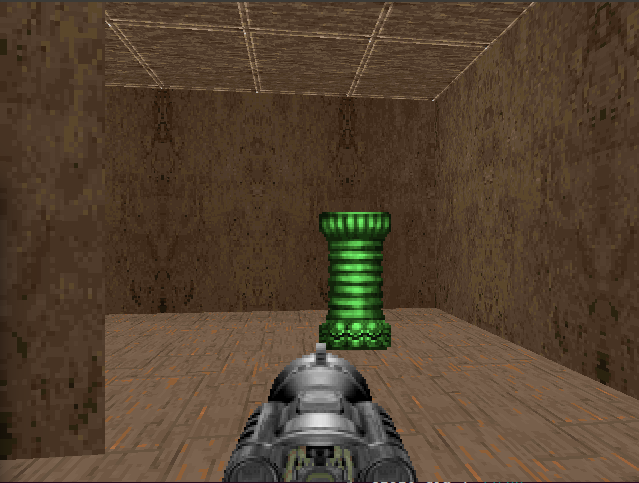}}}$
    \caption{Green Torch $\rightarrow$ Green Tower}
  \end{subfigure} \\
  \vspace{0.1in}
  \begin{subfigure}{0.7\linewidth}
    \centering
    $\vcenter{\hbox{\includegraphics[width=0.33\linewidth]{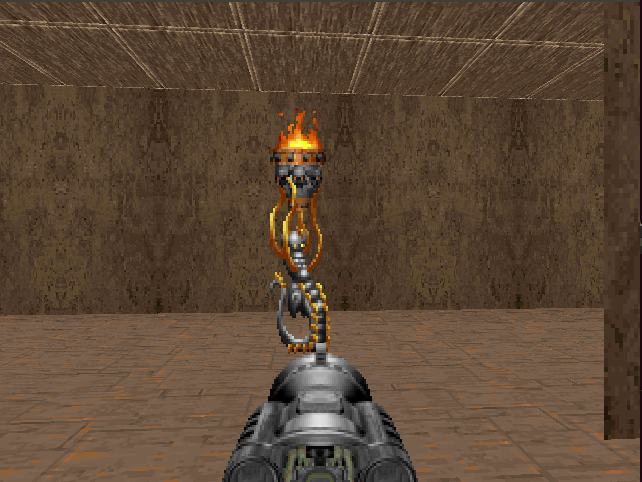}}}$ %
    $\vcenter{\hbox{\scalebox{2}{\Huge\pointer}}}$ %
    $\vcenter{\hbox{\includegraphics[width=0.33\linewidth]{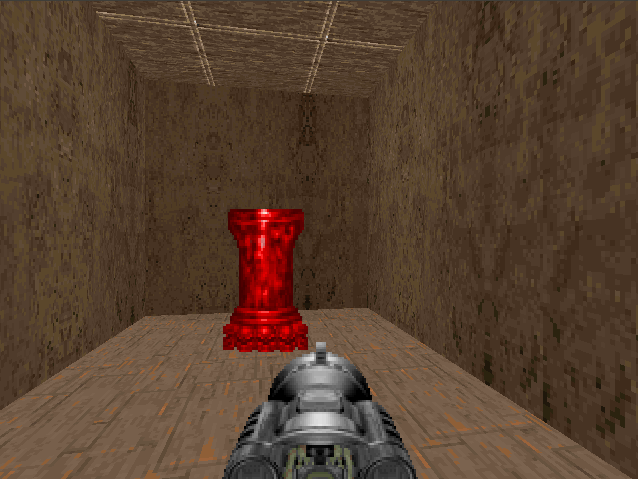}}}$
    \caption{Red Torch $\rightarrow$ Red Tower}
  \end{subfigure}
  \caption{\small Images of the Doom maze environment. The agent starts in the middle of a maze looking in the direction of a torch indicator. The torch can be either green (top-left image) or red (bottom-left image) and indicates which of the goals to search for. The goals are two towers which are randomly located within the maze and match the indicator color. The episode ends whenever the agent touches a tower, whereupon it receives a positive reward if it reached the correct tower, while a negative reward otherwise. Alternatively, the episode is also terminated if the agent has not reached a tower in 420 steps.}
  \label{fig:doomdemo}
\end{figure}

To demonstrate that our method can work in much more complicated 3D environments with longer time lags, we implemented the 2D maze environment in 3D using the ViZDoom~\citep{Kempka2016ViZDoom} environment and a random maze generator. In the Doom environment, the indicator is a torch of either red or green color that is always at a fixed location in view of the player's starting state. The goals are red/green towers that are randomly positioned throughout the maze.
The corresponding torch indicators and goal towers are illustrated in Figure~\ref{fig:doomdemo}. We terminate an episode if more than 420 steps are taken.

To train the agent, we used a deep network as a state embedding. The state observations were the 5 previous $100\times75$ RGB colour images. The network  was pre-initialized with the weights from a network trained to play Doom taken from \cite{DeepDoom16}. We used an action repeat of 5.
To map from 3D positions to neural map space, we rescaled the coordinates provided from the game and did nearest neighbor quantization. The mazes consisted of $3\times3$ rooms, where a wall could be generated between any adjacent room and a tower can be located at the center.
For the Neural Map agent, we used an internal map size of $32\times9\times9$.

The agents we tested on are an LSTM baseline and an LSTM + Neural Map with GRU-based updates.
Due to the large size of the state embedding network, the memory network implementation quickly ran out of memory with more than 16 states so we only trained an agent with a memory size of 16.
Additionally, it was observed that because the Neural Map has coarse granularity in the 3D maze (several positions in the maze occupy the same ``pixel'' position in the map), the neural map agents that lacked an LSTM added to the top often repeated the same actions. So, for example, when the agent would face a wall it would often start turning but the next frame it did not remember which direction it started turning towards. Therefore the agent would sometimes be stuck facing the wall and oscillate between turning left or right. 
To get around this, the LSTM pre-output layer keeps track of what actions were done in previous frames and so can enable the agent to more easily turn in a consistent direction. Due to this behaviour, we do not present results for stand-alone Neural Map agents.

\begin{figure}
  \centering
  \includegraphics[width=0.22\linewidth]{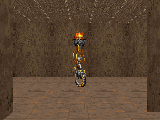}
  \includegraphics[width=0.22\linewidth]{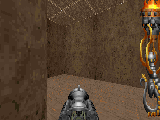}
  \includegraphics[width=0.22\linewidth]{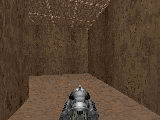}
  \includegraphics[width=0.22\linewidth]{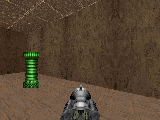} \\
  \includegraphics[width=0.22\linewidth]{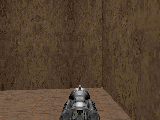}
  \includegraphics[width=0.22\linewidth]{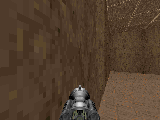}
  \includegraphics[width=0.22\linewidth]{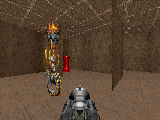}
  \includegraphics[width=0.22\linewidth]{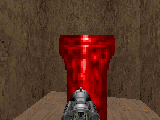}
  \caption{\small An episode showing the LSTM + Neural Map (GRU) agent walking down a corridor, seeing the wrong green goal tower, and then backtracking back to the indicator until it sees the correct red goal tower. Read using left-right, top-bottom order. This episode lasted around 110 time steps, demonstrating that the environment requires information about the indicator to be stored for very long time lags. The MemNN agent would be unable to solve such a maze due to its limited memory size.}
  \label{fig:doomep}
  \vspace{-0.1in}
\end{figure}

We trained all agents on a single training maze (i.e. the wall geometry was constant for all training episodes) for up to 7 million frames. For testing, we used a held-out set of 6 mazes to see whether the agents were capable of zero-shot learning. In both training and testing settings, for every episode we sample new random goal tower locations.  The results are shown in Table~\ref{tab:doomresults}. For each agent, we chose the best performing network that was seen during the 7 million training frames, where performance was measured with respect to the training map.  We can see LSTM + Neural Map (GRU) surpasses all other methods on both the training map and on the 6 unseen maps. On the training map, the LSTM does almost as well as the memory network which is limited to the past 16 frames. This suggests that the LSTM is potentially only learning to solve the scenarios where the goal towers are closer to the indicator. Figure~\ref{fig:doomep} shows an example episode where the Neural Map agent successfully backtracks.

\begin{table}
  \centering
  \begin{tabular}{|c|c|c|}
    \hline
    Agent  & Training Map & Unseen Maps \\ \hline
    Random & 20.9\% & 22.1\% \\ \hline
    MemNN  & 68.2\% & 60.3\% \\ \hline
    LSTM   & 69.2\% & 52.4\% \\ \hline
    \makecell{LSTM+Neural \\ Map (GRU)} & {\bf 78.3\%} & {\bf 66.6\%} \\ \hline
  \end{tabular}
  \caption{\small Doom results showing the percentage of 1000 episodes that resulted in the agent successfully finding the correct goal within 420 steps. We trained all agents on a single map. On one column, we report the success on the training map while on the other we report results on a heldout set of 6 unseen mazes. The goal tower locations are randomly sampled each episode.}
  \label{tab:doomresults}
  \vspace{-0.1in}
\end{table}

\vspace{-0.1in}
\section{Extension: Ego-centric Neural Map}

A major disadvantage of the neural map as previously described is that it requires some oracle to provide the current $(x,y)$ position of the agent. This is a difficult problem in and of itself, and, despite being well studied, it is far from solved. The alternative to using absolute positions within the map is to use relative positions. That is, whenever the agent moves between time steps with some velocity $(u,v)$, the map is counter-transformed by $(-u,-v)$, i.e. each feature in the map is shifted in the $H$ and $W$ dimensions. This will mean that the map will be ego-centric, i.e. the agent's position will stay stationary in the center of the neural map while the world as defined by the map moves around them. Therefore in this setup we only need some way of extracting the agent's velocity, which is typically a simpler task in real environments. Here we assume that there is some function $\xi(u',v')$ that discretizes the agent velocities $(u',v')$ so that they represent valid velocities within the neural map $(u,v)$. In the sequel, we assume that all velocities have been properly normalized by $\xi$ into neural map space.

Let $(pw,ph)$ be the center position of the neural map. The ego-centric neural map operations are shown below:
\begin{align}
  \overline{M}_t              &= CounterTransform(M_t, (u_t,v_t)) \nonumber \\
  r_t                         &= read(\overline{M}_t), \hspace{0.05in} c_t = context( \overline{M}_t, s_t, r_t)  \nonumber \\
  w_{t+1}^{(pw,ph)}           &= write(s_t, r_t, c_t, \overline{M}_t^{(pw,ph)}) \nonumber \\
  M_{t+1}                     &= egoupdate( \overline{M}_t, w_{t+1}^{(pw,ph)}) \nonumber \\
  o_t                         &= [r_t, c_t, w_{t+1}^{(pw,ph)}], \nonumber \\
 \pi_t &= \text{Softmax}(f(o_t))   
\end{align}
where $\overline{M}_t$ is the current neural map reverse transformed by the current velocity $(u_t,v_t)$ so that the agents map position remains in the center $(pw,ph)$. 

{\bf Counter Transform Operation:} \\
The $CounterTransform$ operation transforms the current neural map $M_t$ by the inverse of the agent's current velocity $(u_t, v_t)$. Written formally:
\begin{align}
  \overline{M}_t^{(a,b)} = \left\{ \begin{array}{lr}
    M_{t+1}^{(a-u,b-v)}, & \text{for } \substack{(a-u)\in\{1,\hdots,W\}\wedge \\ (b-v)\in\{1,\hdots,H\}} \\
    0,                   & \text{else}
  \end{array}\right.
\end{align}
While here we only deal with reverse translation, it is possible to handle rotations as well if the agent can measure it's angular velocity.

{\bf Map Egoupdate Operation:} \\
The $egoupdate$ operation is functionally equivalent to the $update$ operation except only the center position $(pw,ph)$ is only ever written to:
\begin{align}
  M_{t+1}^{(a,b)} = \left\{ \begin{array}{lr}
    w_{t+1}^{(pw,ph)}, & \text{for } (a,b)=(pw,ph) \\
    \overline{M}_t^{(a,b)},      & \text{for } (a,b)\neq(pw,ph)
  \end{array}\right.
\end{align}

\vspace{-0.3in}
\section{Related Work}

Other than the straightforward architectures of combining an LSTM with Deep Reinforcement Learning (DRL)~\citep{A3C16, hausknecht2015deep}, there has also been work on using more advanced external memory systems with DRL agents to handle partial observability. \cite{MinecraftMemnet16} used a memory network (MemNN) to solve maze-based environments similar to the ones presented in this paper. MemNN keeps the last $M$ states in memory and encodes them into (key, value) feature pairs.
It then queries this memory using a soft attention mechanism similar to the context operation of the Neural Map, except in the Neural Map the key/value features were written by the agent and aren't just a stored representation of the last $M$ frames seen.
\cite{MinecraftMemnet16} tested a few variants of this basic model, including ones which combined both LSTM and memory-network style memories.

In contrast to memory networks, another research direction is to design recurrent architectures that mimic computer memory systems. These architectures explicitly separate computation and memory in a way anagolous to a modern digital computer, in which some neural controller (akin to a CPU) interacts with an external memory (RAM).
One recent model is similar to the Neural Map, called the Differentiable Neural Computer (DNC)~\citep{DNC16}, which combines a recurrent controller with an external memory system that allows several types of read/write access. In addition to defining an unconstrained write operator (in contrast to the neural map's write location being fixed), the DNC has a selective read operation that reads out the memory either by content or in the order that it was written. While the DNC is more specialized to solving algorithmic problems, the Neural Map can be seen as an extension of this Neural Computer framework to 3D environments, with a specific inductive bias on its write operator that allows sparse writes. Recently work has also been done toward sparsifying the read and write operations of the DNC~\citep{sparseDNC16}.
This work was not focused on 3D environments and did not make any use of task-specific biases like agent location, but instead used more general biases like ``Least-Recently-Used'' memory addresses to force sparsity. 

Additionally, a recent paper has used the idea of augmenting the state of an agent with an internal map when acting in 3D environments~\citep{DoomSlam16}. Their approach uses a sophisticated pipeline of hard-coded sub-modules, such as SLAM (Simultaneous Localization And Mapping), image segmentation, etc., to augment the image inputs that are typically fed to DRL agents.
In contrast, the Neural Map is trained fully end-to-end without even weak supervision and therefore it can learn by itself what currently relevant information it should store within in its internal knowledge map of the environment.

A similar paper that also had a 2D map structured memory was recently made public concurrently with our submission. \cite{CogMap17} designed a spatial memory that was used to do robot navigation in 3D environments. These environments were based off image scans of real office buildings, and they were preprocessed into a grid-world by quantizing the possible positions and orientations the agent could assume. 
In contrast to our paper, which presents the Neural Map more as a general memory architecture for DRL agents, \cite{CogMap17} focuses mainly on solving the task of robot navigation.
More concretely, the task in these environments was to navigate to a goal state, with the goal position either stated semantically (find a chair) or stated in terms of the position relative to the robot's coordinate frame. 
Owing to this focus on navigation, they force their internal map representation (e.g. $M_t$) to be a prediction of free space around the robot.
Finally, their method used DAGGER~\citep{ross2011reduction}, an imitation learning algorithm, to train their agent. Since Doom actions affect translational/rotational accelerations, training using imitation learning is more difficult since a search algorithm cannot be used as supervision. 
An interesting addition they made was the use of a multi-scale map representation and a Value Iteration network~\citep{tamar2016value} to do better path planning.

\section{Conclusion}

In this paper we developed a neural memory architecture that organizes the spatial structure of its memory in the form of a 2D map, and allows sparse writes to this memory where the memory address of the write is in a correspondence to the agent's current position in the environment. We showed its ability to learn, using a reinforcement signal, how to behave within a challenging 2D maze task that required storing information over long time steps. The results demonstrated that our architecture surpassed baseline memories used in previous work. They also revealed that the GRU-based update equation we defined was crucial to improving both learning speed and training stability. Finally, to show that our method can scale up to more difficult 3D environments, we reimplemented the maze environment in Doom. Using a hybrid Neural Map + LSTM model, we were able to solve most of the scenarios, surpassing both LSTM and MemNN baseline agents.



\bibliography{iclr2017_conference}
\bibliographystyle{iclr2017_conference}

\end{document}